\documentclass{article}
\usepackage[dvipdfm]{graphicx}  % platex->dvipdfmx
\usepackage[fleqn]{amsmath}
\usepackage{amssymb}
\usepackage{amsthm}
\usepackage{latexsym}
\usepackage{url}
\def\u{\boldsymbol{u}}

\def\w{\boldsymbol{w}}
\def\x{\boldsymbol{x}}
\def\y{\boldsymbol{y}}

\def\A{\boldsymbol{A}}
\def\B{\boldsymbol{B}}
\def\C{\boldsymbol{C}}
\def\D{\boldsymbol{D}}

\def\I{\boldsymbol{I}}
\def\K{\boldsymbol{K}}
\def\L{\boldsymbol{L}}

\def\QQ{\boldsymbol{Q}}
\def\R{\boldsymbol{R}}
\def\S{\boldsymbol{S}}

\def\U{\boldsymbol{U}}
\def\V{\boldsymbol{V}}
\def\W{\boldsymbol{W}}
\def\X{\boldsymbol{X}}
\def\Y{\boldsymbol{Y}}
\def\Z{\boldsymbol{Z}}
\def\0{\boldsymbol{0}}
\def\1{\boldsymbol{1}}
\def\balpha{\boldsymbol{\alpha}}
\def\bbeta{\boldsymbol{\beta}}

\def\cR{\mathcal R}

\def\ov{\overline}
\def\ul{\underline}
\def\wh{\widehat}

\begin{document}

\title{%
  Designing various multivariate analysis at will\\
  via generalized pairwise expression
}
\author{Akisato Kimura \thanks{NTT Communication Science Laboratories, NTT Corporation, 2-4 Hikari-dai, Seika, Soraku, Kyoto, 619--0237 Japan.} \and Masashi Sugiyama \thanks{Graduate School of Information Science and Engineering, Tokyo Institute of Technology, 2-12-1 Oookayama, Meguro, Tokyo, 152--8552 Japan.} \and Hitoshi Sakano \thanks{NTT Communication Science Laboratories, NTT Corporation, 2-4 Hikari-dai, Seika, Soraku, Kyoto, 619--0237 Japan.} \and Hirokazu Kameoka \thanks{Graduate School of Information Science and Technologies, the University of Tokyo, 7-3-1 Hongo, Bunkyo, Tokyo, 113--8656 Japan.}}

\begin{abstract}
It is well known that dimensionality reduction based on multivariate analysis methods and their kernelized extensions can be formulated as generalized eigenvalue problems of scatter matrices, Gram matrices or their augmented matrices.
This paper provides a generic and theoretical framework of multivariate analysis introducing a new expression for scatter matrices and Gram matrices, called Generalized Pairwise Expression (GPE).
This expression is quite compact but highly powerful.
The framework includes not only (1) the traditional multivariate analysis methods but also (2) several regularization techniques, (3) localization techniques, (4) clustering methods based on generalized eigenvalue problems, and (5) their semi-supervised extensions.
This paper also presents a methodology for designing a desired multivariate analysis method from the proposed framework.
The methodology is quite simple: adopting the above mentioned special cases as templates, and generating a new method by combining these templates appropriately.
Through this methodology, we can freely design various tailor-made methods for specific purposes or domains.
\end{abstract}

%\begin{keyword}
%Multivariate analysis, dimensionality reduction, generalized eigenvalue problem, %pairwise expression, kernel method, clustering, semi-supervised learning, %regularization
%\end{keyword}

\allowdisplaybreaks  % allow to repage in the mathematical environments
\maketitle

%%%%%%%%
\section{Introduction}
\label{sec:intro}

%\begin{figure}[t]
%  \begin{center}
%    \includegraphics[width=0.975\hsize]{overview1.png}\\ \vspace{5mm}
%    \includegraphics[width=0.975\hsize]{overview2.png}
%  \end{center}
%  \caption{Overview of the proposed method \label{fig:overview} }
%\end{figure}

We can easily obtain a massive collection of texts (long articles\footnote{New York Times Article Archive: \url{http://www.nytimes.com/ref/membercenter/nytarchive.html}}, microblogs\footnote{Tweets2011 corpus for TREC2011 microblog track: \url{http://trec.nist.gov/data/tweets/}}), images \cite{labelMe,imageNet,tinyImages,arista}, videos \cite{TRECVID:conf,labelMeVideo} and musics \cite{millionsong}\footnote{Last.fm: \url{http://www.lastfm.jp}, Freesound: \url{http://www.freesound.org}} nowadays.
However, we are now facing a difficulty in finding an intrinsic trend and nature of such a massive collection of data.
Multivariate analysis \cite{multivariate:Anderson} is traditional, quite simple but might be one of the powerful tools to obtain a hidden structure embedded in the data, via classification, regression and clustering \cite{PRML,Hastie}.
Actually, multivariate analysis has been still an important tool, and recent reports showed its effectiveness for several applications, e.g. human detection \cite{humanDetectionWithPLS}, image annotation \cite{dimensionReductionForImageAnnotation,correlationalSpectralClustering}, sensor data mining \cite{fda_sensors,cca_sensors2,cca_sensors3,cca_sensors4}.
Principal component analysis (PCA) \cite{CCA:hotelling}, Fisher discriminant analysis (FDA) \cite{FDA}, multivariate linear regression (MLR), canonical correlation analysis (CCA) \cite{CCA:hotelling}, and partial least squares (PLS) \cite{PLS} are well known as standard multivariate analysis methods.
These methods can be formulated as a generalized eigenvalue problem of a scatter matrix or an augmented matrix composed of several scatter matrices.
Several extended researches tried to tackle the small sample size problem \cite{smallSampleSize}, i.e., the situation where the number of training samples is small compared with their dimensionality (e.g. robust PCA \cite{robustPCAFirst,robustPCA,robustSubspace,robustMultiLinearPCA} and robust FDA \cite{robustFDA,subclassFDA,mixtureSubclassFDA}).
Kernelized extensions of those standard methods have been also developed to deal with non-vector samples and non-linear analysis (e.g. kernel PCA \cite{kernelPCA}, kernel FDA \cite{kernelFDA1,kernelFDA2,extendedKernelFDA}, kernel MLR \cite{kernelMLR_bishop}, kernel CCA \cite{kernelCCA:akaho,kernelCCA:fyfe}).
They can be formulated as a generalized eigenvalue problem of an augmented matrix composed of Gram matrices, instead of scatter matrices.
Kernel multivariate analysis often needs some regularization techniques such as $\ell_2$-norm regularization \cite{ridgeRegression,tikhonov,NC:Hardoon:2004} to inhibit overfitting and graph Laplacian method \cite{semiSupervisedKCCA:blaschko} to fit underlying data manifolds smoothly.
In addition, improvements of robustness against outliers and non-Gaussianity (i.e. multi-dimensional scaling (MDS) \cite{MDS}, locality preserving projection (LPP) \cite{LPP} and local Fisher discriminant analysis (LFDA) \cite{LFDA_paper}) and their extensions to semi-supervised dimensionality reduction \cite{SELF_paper,semiSupervisedKCCA:blaschko,SemiCCA_icpr} have been considered.
In addition, a lot of multivariate analysis methods and several trials to unify these methods have been presented so far.
Borge et al \cite{Borga97unified} and De Bie et al \cite{DeBie05unified} showed that several major linear multivariate analysis method can be formulated by a unified form of generalized eigenvalue problems by introducing the augmented matrix expression.
Sun et al \cite{leastSquareFormulation:Sun,leastSquareFormulationPAMI:Sun} showed the equivalence between a certain class of generalized eigenvalue problems and least squares problems under a mild assumption.
De la Torre \cite{DelaTorre2010,ls-wkrrrPAMI} further extended their work to a various kind of component analysis methods by introducing the formulation of least-squares weighted kernel reduced rank regression (LS-WKRRR).
However, freely designing a tailor-made multivariate analysis for a specific purpose or domain still remains an open problem.
Until now, researchers and engineers have had to choose one of the existing methods that seems best to address the problem of interest, or had to laboriously develop a new analysis method tailored specifically for that purpose.
In view of the above discussions, this paper provides a new expression of
covariance matrices and Gram matrices, which we call the \emph{Generalized Pairwise Expression (GPE)} to make it easy to design a new multivariate analysis method with desired property.
The methodology is quite simple:
Exploiting the above mentioned existing methods as templates, and constructing a new method by combining these templates appropriately.
This characteristics has not been discussed yet in any previous researches to our best knowledge.
It is also possible to individually select and arrange samples for calculating the scatter matrices of the methods to be combined, which enables us to extend multivariate analysis methods to semi-supervised ones and multi-modal ones, where some parts are calculated from only labeled samples, and the other parts are obtained from both labeled and unlabeled samples.
The rest of this paper is organized as follows:
Section \ref{sec:linear} defines a class of multivariate analysis methods we are concerned with in this paper.
Next, Section \ref{sec:gpe} describes our proposed framework, GPE, and its fundamental properties.
These properties provide a methodology to design multivariate analysis methods with desired properties.
Then, Section \ref{sec:special} reviews major multivariate analysis methods from the viewpoint of GPE.
This review will give us templates of the GPEs for designing desired methods.
After the above preparations, Section \ref{sec:new} demonstrates how to design a new multivariate analysis method.
By replicating the methodology shown in the preceding sections, we can easily design various multivariate analysis methods at will.
Additionally, Section \ref{sec:kernel} describes a non-linear and/or non-vector extension of GPE with the help of the kernel trick.
With this extension, non-linear dimensionality reduction, and several clustering methods are all in the class of multivariate analysis methods we are concerned with.
%

%%%%%%%%
\section{Multivariate analysis for vector data}
\label{sec:linear}

Consider two sets $\X$ and $\Y$ of samples\footnote{The following discussion
can be easily extended to more than 2 sets of samples sets \cite{generalizedCCA:yanai}.}, where each set contains $N_x$ and $N_y$ samples, and each sample can be expressed as a vector with $d_x$ and $d_y$ dimensions, respectively, as follows:
\begin{eqnarray*}
  \X &=& \{\x_1,\ldots,\x_{N_x}\},\\
  \Y &=& \{\y_1,\ldots,\y_N,\y_{N_x+1},\ldots,\y_{N_x+N_y-N}\}.
\end{eqnarray*}
For brevity, both of the sample sets $\X$ and $\Y$ are supposed to be centered on the origin by subtracting the mean from each component.
Suppose that samples $\x_n$ and $\y_n$ with the same suffix are co-occurring.
Each set $\X$ and $\Y$ of samples is separated into the following two types:
{\it Complete sample sets} $\X^{({\rm C})}$ and $\Y^{({\rm C})}$ so that every sample $\x_n$ (resp. $\y_n$) has co-occurring sample $y_n$ (resp. $\x_n$), and {\it incomplete sample sets} $\X^{({\rm I})}$ and $\Y^{({\rm I})}$ so that every sample $\x_n$ (resp. $\y_n$) cannot find the co-occurring sample.
\begin{eqnarray*}
  \X^{({\rm C})} &=& \{\x_1,\x_2,\ldots,\x_N\},\\
       &=& \{\x_1^{({\rm C})},\x_2^{({\rm C})},\ldots,\x_N^{({\rm C})}\},\\
  \Y^{({\rm C})} &=& \{\y_1,\y_2,\ldots,\y_N\},\\
       &=& \{\y_1^{({\rm C})},\y_2^{({\rm C})},\ldots,\y_N^{({\rm C})}\},\\
  \X^{({\rm I})} &=& \{\x_{N+1},\x_{N+2},\ldots,\x_{N_x}\},\\
       &=& \{\x_1^{({\rm I})},\x_2^{({\rm I})},\ldots,\x_{N_x-N}^{({\rm I})}\},\\
  \Y^{({\rm I})} &=& \{\y_{N_x+1},\y_{N_x+2},\ldots,\y_{N_x+N_y-N}\},\\
       &=& \{\y_1^{({\rm I})},\y_2^{({\rm I})},\ldots,\y_{N_y-N}^{({\rm I})}\}
\end{eqnarray*}
First, we concentrate on the case that $N_x=N_y=N$, namely all the samples are
paired, unless otherwise stated.
Many linear multivariate analysis methods developed so far involve an optimization problem of the following form:
\begin{eqnarray}
  \w^{\rm (opt)} &=& \arg\max_{\w\in\cR^d}R(\w),\label{eq:linear1}\\
  R(\w) &=& \w^{\top}\ov{\C}\w(\w^{\top}\ul{\C}\w)^{-1},\nonumber
\end{eqnarray}
where $\ov{\C}$ and $\ul{\C}$ are square matrices with certain statistical nature.
For example, $\ov{\C}$ is a scatter matrix of $\X$ and $\ul{\C}$ is an identity
matrix in PCA, and $\ov{\C}$ is a between-class scatter matrix and $\ul{\C}$ is
a within-class scatter matrix in FDA.
Roughly speaking, $\ov{\C}$ encodes the quantity that we want to increase, and
$\ul{\C}$ corresponds to the quantity that we want to decrease.
The denominator of the function $R(\w)$ is often normalized to remove scale ambiguity, resulting in the following form:
\begin{eqnarray}
  \w^{\rm (opt)} &=& \arg\max_{\w\in\cR^d} R_1(\w)~
    \mbox{s.t. } R_2(\w)=1,\label{eq:linear2}\\
  R_1(\w) &=& \w^{\top}\ov{\C}\w,\quad R_2(\w) = \w^{\top}\ul{\C}\w.
    \nonumber
\end{eqnarray}
The above optimization problem can be converted to the following generalized
eigenvalue problem via the Lagrange multiplier method:
\begin{eqnarray}
  \ov{\C}\w &=& \lambda\ul{\C}\w. \label{eq:linear3}
\end{eqnarray}
The solution $\w_k$ $(k=1,2,\ldots,r)$ of the above generalized eigenvalue
problem gives a solution of the original multivariate analysis formulated in
Equation (\ref{eq:linear1}).
It can be confirmed that Equation (\ref{eq:linear1}) is invariant against any kinds of linear transformations, i.e., a vector $\U\w^{\rm (opt)}$ transformed by any $r$-dimensional unitary matrix $\U$ is also a global solution.
This implies that the range of the embedding space can be uniquely determined by Equation (\ref{eq:linear1}), but the metric in the embedding space is arbitrary.
A practically useful heuristic is to set
\begin{eqnarray}
  \U &=& \mbox{diag}(\sqrt{\lambda_1},\sqrt{\lambda_2},\ldots,\sqrt{\lambda_d}),
\end{eqnarray}
where $\mbox{diag}(a,b,\cdots,c)$ denotes the diagonal matrix with the diagonal elements $a,b,\ldots,c$, and $\{\lambda_k\}_{k=1}^r$ denotes the generalized eigenvalues.
Finally, we obtain the solution as
\begin{eqnarray}
  \W^{\rm (opt)}
  &=& \{\sqrt{\lambda_1}\w_1,\sqrt{\lambda_2}\w_2,\ldots,\sqrt{\lambda_r}\w_r\}.
\end{eqnarray}
Thus, the minor eigenvectors are de-emphasized according to the square root of the eigenvalues.
%

%%%%%%%%
\section{Generalized pairwise expression}
\label{sec:gpe}

%\begin{figure}[t]
%  \begin{center}
%    \includegraphics[width=0.975\hsize]{gpe.png}
%  \end{center}
%  \caption{
%    Various multivariate analysis methods can be described via generalized %pairwise expression (GPE)
%    \label{fig:gpe}
%  }
%\end{figure}

When addressing linear multivariate analysis methods, we often deal with the following type of second-order statistics as an extension of scatter matrices, since it is convenient to describe the relation between two features regarding whether they are close together or far apart
\begin{eqnarray*}
  \S_{\QQ,xy}
  &=& \sum_{n=1}^{N}\sum_{m=1}^{N} Q_{n,m}(\x_n-\x_m)(\y_n-\y_m)^{\top},
\end{eqnarray*}
where $\QQ$ is an $N\times N$ non-negative, semi-definite and symmetric matrix.
A typical example is the scatter matrix\footnote{Due to the limited space, we
describe only the scatter matrix $\S_{xy}$ and its extensions with the pairwise
form. The scatter matrices $\S_{xx}$ and $\S_{yy}$, and their extensions can
be easily derived in the same way.}:
\begin{eqnarray*}
  \S_{xy} &=& {\textstyle N^{-1}\sum_{n=1}^N \x_n\y_n^{\top}}.
\end{eqnarray*}
Let $\D_{\QQ}$ be the $N\times N$ diagonal matrix with
\begin{eqnarray*}
  D_{\QQ,n,n} &=& \sum_{n_2=1}^{N} Q_{n,n_2},
\end{eqnarray*}
and let $\L_{\QQ}$ be $\L_{\QQ}=\D_{\QQ}-\QQ$.
Then, the matrix $\S_{\QQ,xy}$ can be expressed in terms of $\L_{\QQ}$ as follows:
\begin{eqnarray*}
  \S_{\QQ,xy} &=& \X\L_{\QQ}\Y^{\top}.
\end{eqnarray*}
The above expression is called the \emph{pairwise expression (PE)} of the second-order statistics $\S_{\QQ,xx}$\cite{SELF_paper}.
If $\QQ$ is a weight matrix for a graph with $n$ nodes, $\L_{\QQ}$ can be regarded as a graph Laplacian matrix in the spectral graph theory.
If $\QQ$ is symmetric and its elements are all non-negative, $\L_{\QQ}$ is known to be positive semi-definite.

Here, we extend PE to the following expression introducing an additional matrix
independent of $\QQ$:
\begin{eqnarray*}
  \hat{\S}_{\QQ,xy} &=& \X\L_{\QQ,1}\Y^{\top}+\L_2,
\end{eqnarray*}
where $\L_{\QQ,1}$ is a $N\times N$ positive semi-definite matrix, and $\L_2$
is a $d_x\times d_y$ non-negative semi-definite matrix.
We do not have to explicitly consider the matrix $\QQ$ for the following discussions:
\begin{eqnarray}
  \hat{\S}_{xy} &=& \X\L_1\Y^{\top}+\L_2. \label{eq:general:pairwise1}
\end{eqnarray}
After all, we call this expression as the \emph{generalized pairwise expression
(GPE)}.
The first term of Equation (\ref{eq:general:pairwise1}) is called the \emph{data term} since it depends on the sample data, and the second term is called the \emph{bias term}.
We can derive the following fundamental properties of GPE from the definition, if the number of samples, $N$ is sufficiently large:
\begin{enumerate}
\item
  If $\A$ is GPE and $\beta>0$ is a constant, then $\beta\A$ is also GPE.
\item
  If both $\A$ and $\B$ are GPE with $d_x$ rows and $d_y$ columns, then $\A+\B$ is also GPE with $d_x$ rows and $d_y$ columns.
\item
  If $\A$ is GPE with $d_x$ rows and $d_y$ columns, and $\B$ is GPE with $d_y$ rows and $d_z$ columns, then $\A\B$ is also GPE with $d_x$ rows and $d_z$ columns.
\end{enumerate}

\begin{proof}
The first and second claims can be easily proved, so we concentrate on proving the third one.
First, let us denote $\A$ and $\B$ as follows:
\begin{eqnarray*}
  \A &=& \X\L_{A1}\Y^{\top}+\L_{A2},\\
  \B &=& \Y\L_{B1}\Z^{\top}+\L_{B2},
\end{eqnarray*}
where $\L_{A1}$ (resp. $\L_{B1}$) is a positive semi-definite matrix with $d_x$ (resp. $d_y$) rows and $d_y$ (resp. $d_z$) columns, and $\L_{A2}$ (resp. $\L_{B2}$) is a $d_x\times d_y$ (resp. $d_y\times d_z$) non-negative matrix.
Then, we obtain
\begin{eqnarray*}
  \A\B
  &=& (\X\L_{A1}\Y^{\top}+\L_{A2})(\Y\L_{B1}\Z^{\top}+\L_{B2}),\\
  &=& \X(\L_{A1}\Y^{\top}\Y\L_{B1})\Z^{\top}+(\L_{A2}\Y)\L_{B1}\Z^{\top}\\
  & & +\X\L_{A1}(\Y^{\top}\L_{B2})+\L_{A2}\L_{B2}.
\end{eqnarray*}
Here, we can find some matrices $\L_{Ci}$ ($i=1,2,3$) satisfying the following relationships, if $N\ge\max(d_x,d_y,d_z)$:
\begin{eqnarray*}
  \L_{C1}   &=& \L_{A1}\Y^{\top}\Y\L_{B1},\\
  \X\L_{C2} &=& \L_{A2}\Y,\\
  \L_{C3}\Z^{\top} &=& \Y^{\top}\L_{B2}.
\end{eqnarray*}
This implies that
\begin{eqnarray*}
  \lefteqn{\A\B}\nonumber\\
  &=& \X\L_{C1}\Z^{\top}+\X\L_{C2}\L_{B1}\Z^{\top}\\
  & & \hspace{10mm}+\X\L_{A1}\L_{C3}\Z^{\top}+\L_{A2}\L_{B2}\\
  &=& \X(\L_{C1}+\L_{C2}\L_{B1}+\L_{A1}\L_{C3})\Z^{\top}+\L_{A2}\L_{B2}\\
  &=& \X\L_{D1}\Z^{\top}+\L_{D2},
\end{eqnarray*}
for some matrices $\L_{D1}$ and $\L_{D2}$, which means $\A\B$ is also GPE.
\end{proof}
Recall that the class of multivariate analysis we are dealing with can be expressed as $\ov{\C}\w=\lambda\ul{\C}\w$, and both $\ov{\C}$ and $\ul{\C}$ can be expressed by GPEs or their augmented matrices.
The notable point is that various multivariate analysis methods can be easily designed with the help of these GPE properties, namely by combining GPEs of existing methods with desired properties.
The rest of the problem is to reveal GPE of existing methods and the function of every type of combinations (addition and/or multiplication), which will be described in the next section.
%

%%%%%%%%
\section{Reviewing multivariate analysis}
\label{sec:special}

%%%%
\subsection{Preliminaries}
\label{sec:special:pre}

This section reviews major multivariate analysis methods from the viewpoint of GPE.
As shown in Sections \ref{sec:linear} and \ref{sec:gpe}, the GPEs of PCA and FDA
respectively are given by
\begin{eqnarray*}
  \ov{\C}^{\rm (PCA)} &=& \S_{xx},\quad \ul{\C} ~ = ~ \I_{d_X},\\ 
  \ov{\C}^{\rm (FDA)} &=& \S_{xx}^{\rm (b)},\quad \ul{\C} ~ = ~ \S_{xx}^{\rm (w)},
\end{eqnarray*}
where $\S_{xx}^{\rm (b)}$ and $\S_{xx}^{\rm (w)}$ are respectively between-class and within-class scatter matrices of $\X$.
From these examples, a scatter matrix $\S_{xx}$ is a typical example of the data term in GPE, and an identity matrix $\I_{d}$ is a typical example of the bias term.
%

%%%%
\subsection{Canonical correlation analysis (CCA)}
\label{sec:special:CCA}

Canonical correlation analysis (CCA)\cite{CCA:hotelling} is a method of correlating linear relationships between two sample sets.
Formally, CCA finds a new coordinate $(\w_x,\w_y)$ to maximize the correlation between the two vectors in the new coordinates.
In other words, the function $\rho(\w_x,\w_y|\X,\Y)$ to be maximized is
\begin{eqnarray}
  \lefteqn{\rho^{\rm (CCA)}(\w_x,\w_y|\X,\Y)}\nonumber\\
  &=& \frac{\langle\X^{\top}\w_x,\Y^{\top}\w_y\rangle}
      {\|\X^{\top}\w_x\|\cdot\|\Y^{\top}\w_y\|}\nonumber\\
  &=& \max_{(\w_x,\w_y)}
      \frac{\wh{E}[\langle\w_x,\x\rangle\langle\w_y,\y\rangle]}
      {\sqrt{\wh{E}[\langle\w_x,\x\rangle^2]\cdot
      \wh{E}[\langle\w_y,\y\rangle^2]}}\nonumber\\
  &=& \max_{(\w_x,\w_y)}\frac{\w_x^{\top}\wh{E}[\x\y^{\top}]\w_y}
      {\sqrt{\w_x^{\top}\wh{E}[\x\x^{\top}]\w_x\w_y^{\top}
      \wh{E}[\y\y^{\top}]\w_y}}\nonumber\\
  &=& \frac{\w_x^{\top} \S_{xy} \w_y}
      {\sqrt{\w_x^{\top} \S_{xx}\w_x\w_y^{\top} \S_{yy}\w_y}}.
      \label{eq:special:CCA1}
\end{eqnarray}
The maximum of the function $\rho(\X^{({\rm C})},\Y^{({\rm C})})$ is not affected by re-scaling $\w_x$ and $\w_y$ either together or independently.
Therefore, the maximization of $\rho(\X^{({\rm C})},\Y^{({\rm C})})$ is equivalent to maximizing the numerator of $\rho(\X^{({\rm C})},\Y^{({\rm C})})$ subject to
\begin{eqnarray*}
  && \w_x^{\top} \S_{xx} \w_x = \w_y^{\top} \S_{yy} \w_y = 1.
\end{eqnarray*}
Taking derivatives of the corresponding Lagrangian with respect to $\w_x$ and $\w_y$, we obtain
\begin{eqnarray*}
  \S_{xy}\w_y-\lambda \S_{xx}\w_x &=& \0,\\
  \S_{yx}\w_x-\lambda \S_{yy}\w_y &=& \0.
\end{eqnarray*}
From the above discussion, the GPE of CCA can be obtained as follows:
\begin{align*}
  \ov{\C}^{\rm (CCA)} &=
  \begin{pmatrix}
    \0 & \S_{xy}\\
    \S_{yx} & \0\\
  \end{pmatrix},&
  \ul{\C}^{\rm (CCA)} &=
  \begin{pmatrix}
    \S_{xx} & \0\\
    \0 & \S_{yy}\\
  \end{pmatrix}.
\end{align*}
We additionally note that when every sample $\y_n$ in $\Y$ represents a class indicator vectors, namely $\y_n\in\{0,1\}^M$, $\sum_{m=1}^M y_{n,i}=1$ and $M$ is the number of classes, CCA is reduced to FDA \cite{stochasticCCA:bach}\footnote{Note that the technical report \cite{stochasticCCA:bach} includes several mistakes in the discussion as to the equivalence between CCA and FDA.}.
Thus, CCA can be regarded as a generalized variant of FDA so that each sample can belong to multiple classes.
%

%%%%
\subsection{Multiple linear regression (MLR)}
\label{sec:special:MLR}

Multiple linear regression (MLR) is a method of finding a projection matrix $\W$ with the minimum squared error between $\y$ and its linear approximation $\W\x$.
For simplicity, we first consider the case that the projection matrix $\W$ is with rank $1$, which can be written as a direct product of two bases $\w_x$ and $\w_y$.
This assumption is useful to understand MLR from the viewpoint of GPE.
Then, the objective function to be minimized is the following squared error:
\begin{eqnarray*}
  \lefteqn{\epsilon^{\rm (MLR)}(\w_x,\w_y|\X,\Y)}\nonumber\\
  &=& \hat{E}\left[\|\y-\alpha\w_y\w_x^{\top}\x\|^2\right]\\
  &=& \hat{E}\left[\y^{\top}\y\right]
      -2\alpha\w_y^{\top}\hat{E}\left[\y^{\top}\x\right]\w_x 
      +\alpha^2\w_x^{\top}\hat{E}\left[\x^{\top}\x\right]\w_x\\
  &=& \hat{E}\left[\y^{\top}\y\right]-2\alpha\w_y^{\top}\S_{xy}^{\top}\w_x
      +\alpha^2\w_x^{\top}\S_{xx}\w_x,
\end{eqnarray*}
where $\hat{E}[\cdot]$ denotes empirical expectation.
To get an expression for $\alpha$, we calculate the derivative
\begin{eqnarray}
  \lefteqn{\frac{\partial}
                {\partial\alpha}\epsilon^{\rm (MLR)}(\w_x,\w_y|\X,\Y)}\\
  &=& 2(\alpha\w_x^{\top}\S_{xx}\w_x-\w_y^{\top}\S_{xy}^{\top}\w_x)=0,
\end{eqnarray}
which gives
\begin{eqnarray}
  \alpha &=& (\w_y^{\top}\S_{xy}^{\top}\w_x)(\w_x^{\top}\S_{xx}\w_x)^{-1}.
\end{eqnarray}
Then, we obtain
\begin{eqnarray}
  \lefteqn{\epsilon^{\rm (MLR)}(\w_x,\w_y|\X,\Y)}\nonumber\\
  &=& E\left[\y^{\top}\y\right]-
      \frac{(\w_y^{\top}\S_{xy}^{\top}\w_x)^2}{\w_x^{\top}\S_{xx}\w_x}.
      \label{eq:special:MLR1}
\end{eqnarray}

Since the squared error cannot be negative and the first term of objective function is independent of the two directions $\w_x$ and $\w_y$, we can minimize it by maximizing the following generalized Rayleigh quotient:
\begin{eqnarray*}
  \rho^{\rm (MLR)}(\w_x,\w_y|\X,\Y)
  &=& \frac{\w_x^{\top}\S_{xy}\w_y}{\sqrt{\w_x^{\top}\S_{xx}\w_x\w_y^{\top}\w_y}},
\end{eqnarray*}
where $\w_x$ and $\w_y$ are supposed to be normalized as $\w_x^{\top}\S_{xx}\w_x=1$ and $\w_y^{\top}\w_y=1$.
By comparing the above equation and Equation (\ref{eq:special:CCA1}) and the objective function for CCA, we can see that MLR is a special case of CCA, and
\begin{align*}
  \ov{\C}^{\rm (MLR)} &=
    \begin{pmatrix}
      \0 & \S_{xy}\\
      \S_{yx} & \0\\
    \end{pmatrix},&
  \ul{\C}^{\rm (MLR)} &=
    \begin{pmatrix}
      \S_{xx} & \0\\
      \0 & \I_{d_y}\\
    \end{pmatrix}.
\end{align*}
The above derivation shows a part of the equivalence between the generalized eigenproblem and the least squares, which have been already revealed by Sun et al \cite{leastSquareFormulation:Sun,leastSquareFormulationPAMI:Sun}.
This equivalence property will be often exploited in the following discussions.
%

%%%%
\subsection{Principal component regression (PCR)}
\label{sec:special:PCR}

Principal component regression (PCR) \cite{PCR} is a variant of MLR that uses PCA when estimating regression coefficients $\W$.
It is a procedure used to overcome problems which arise when the exploratory variables are nearly co-linear.
In PCR, instead of regressing the dependent variable $\y$ on the independent variables $\x$ directly, the principal components $\V\x$ of the independent variables are used.
One typically only uses a subset of the principal components in the regression, making a kind of regularized estimation.
Often the principal components with the highest variance are selected.
A larger class of multivariate analysis methods that introduces a latent model into the standard linear regression is called as latent variable regression (LVR) \cite{LVR}.
In the same way as MLR, we assume that the projection matrix $\W$ is with rank $1$, namely $\W=\w_y\w_x^{\top}$.
A rank-$K$ approximation $\hat{\X}$ of the data matrix $\X$ can be obtained by singular value decomposition as
\begin{eqnarray}
  \hat{\X} &=& \U_K\Sigma_K\V_K^{\top},
\end{eqnarray}
where $\Sigma_K$ is a $K\times K$ diagonal matrix whose diagonal components are top-$K$ eigenvalues obtained by PCA of $\X$, and $\V_K$ is a $K\times d_x$ matrix whose columns are the top-$k$ eigenvectors.
Then, the objective function of PCR to be minimized can be obtained by substituting $\hat{\X}$ into $\X$ in the objective function of MLR, as follows:
\begin{eqnarray*}
  \lefteqn{\epsilon^{\rm (PCR)}(\w_x,\w_y|\X,\Y)}\nonumber\\
  &=& \epsilon^{\rm (MLR)}(\w_x,\w_y|\hat{\X},\Y)\\
  &=& \hat{E}\left[\|\y-\alpha\w_y\w_x^{\top}\hat{\x}\|^2\right]\\
  &=& \hat{E}\left[\y^{\top}\y\right]
      -2\alpha\w_y^{\top}\hat{E}\left[\y^{\top}\hat{\x}\right]\w_x 
      +\alpha^2\w_x^{\top}\hat{E}\left[\y^{\top}\hat{\x}\right]\w_x\\
  &=& \hat{E}\left[\y^{\top}\y\right]-2\alpha\w_y^{\top}\S_{\hat{x}y}^{\top}\w_x
      +\alpha^2\w_x^{\top}\S_{\hat{\x}\hat{\x}}\w_x,
\end{eqnarray*}
where $\S_{\hat{x}y}$ and $\S_{\hat{x}\hat{x}}$ can be obtained as follows:
\begin{eqnarray*}
  \S_{\hat{x}y}
  &=& \frac{1}{N}\sum_{n=1}^N \u_{K,n}\Sigma_K\V_K^{\top}\y_n^{\top},\\
  \S_{\hat{x}\hat{x}}
  &=& \frac{1}{N}\sum_{n=1}^N
      \u_{K,n}\Sigma_K\V_K^{\top}\V_K\Sigma_K\u_{K,n},\\
  &=& \frac{1}{N}\sum_{n=1}^N \u_{K,n}(\Sigma_K)^2\u_n^{\top}.
\end{eqnarray*}
From the description of the previous subsection, we can obtain
\begin{align*}
  \ov{\C}^{\rm (PCR)} &=
    \begin{pmatrix}
      \0 & \S_{\hat{x}y}\\
      \S_{\hat{x}y}^{\top} & \0\\
    \end{pmatrix},&
  \ul{\C}^{\rm (PCR)} &=
    \begin{pmatrix}
      \S_{\hat{x}\hat{x}} & \0\\
      \0 & \I_{d_y}\\
    \end{pmatrix}.
\end{align*}
%

%%%%
\subsection{Partial Least Squares (PLS)}
\label{sec:special:PLS}

Partial Least Squares (PLS) \cite{PLS} (or sometimes called PLS regression) belongs to a family of latent variable regression (LVR), and tries to finds a direction for the observable sample set $\X$ that explains the maximum variance direction for the predicted sample set $\Y$.
The contribution of PLS against the standard MLR and PCR is to simultaneously estimate the latent model and regression from the latent space to the predicted space, which leads to robust regression against noisy observations.
Although PLS cannot be formulated as a generalized eigenproblem in general, orthogonal PLS (OPLS) \cite{OPLS,OPLS2} as a variant of the original PLS has a form of generalized eigenproblem.
This improves the interpretability (but not the predictivity) of the original PLS.
OPLS can be formulated as follows:
\begin{eqnarray*}
  \X\Y^{\top}\Y\X^{\top}\w &=& \lambda\X\X^{\top}\w,
\end{eqnarray*}
meaning,
\begin{align*}
  \ov{\C}^{\rm (OPLS)}
  &= \X\Y^{\top}\Y\X^{\top}
  &\hspace{-10mm}&\propto \S_{xy}\S_{xy}^{\top},\\
  \ul{\C}^{\rm (OPLS)}
  &= \X\X^{\top}
  &\hspace{-10mm}&\propto \S_{xx}.
\end{align*}
When every sample $\y_n$ in $\Y$ represents a class indicator vectors (cf. Sec \ref{sec:special:CCA}), OPLS is called OPLS-discriminant analysis (OPLS-DA) \cite{OPLS2}, which has been often used for the task of bio-marker identification \cite{OPLS-DA_biomarker}.
%

%%%%
\subsection{$\ell_2$-norm regularization}
\label{sec:special:Tikhonov}

$\ell_2$-norm regularization is a popular regularization technique for various
optimization problems including multivariate analysis.
In the area of statistics or machine learning, this is sometimes called Tikhonov regularization \cite{tikhonov,PRML}.
The most popular method that utilizes $\ell_2$-norm regularization is ridge regression \cite{ridgeRegression}, which combines MLR and $\ell_2$-norm regularization.
The objective function to be minimized is the following squared error:
\begin{eqnarray*}
  \lefteqn{\epsilon^{\rm (Ridge)}(\w_x,\w_y|\X,\Y)}\nonumber\\
  &=& \hat{E}\left[\|\y-\alpha\w_y\w_x^{\top}\x\|^2\right]+\delta\|\w_x\|^2\\
  &=& \hat{E}\left[\y^{\top}\y\right]-2\alpha\w_y^{\top}\S_{xy}^{\top}\w_x
      +\alpha^2\w_x^{\top}\S_{xx}\w_x+\delta\|\w_x\|^2\nonumber\\ \\
  &=& \hat{E}\left[\y^{\top}\y\right]-2\alpha\w_y^{\top}\S_{xy}^{\top}\w_x
      +\alpha^2\w_x^{\top}(\S_{xx}+\hat{\delta}\I_{d_x})\w_x,\nonumber\\
\end{eqnarray*}
where $\hat{\delta}=\delta/\alpha^2$.
From the above equation and the objective function of MLR, the GPE of ridge regression can be derived as
\begin{eqnarray*}
  \ov{\C}^{\rm (Ridge)} &=&
    \begin{pmatrix}
      \0 & \S_{xy}\\
      \S_{yx} & \0\\
    \end{pmatrix},\\
  \ul{\C}^{\rm (Ridge)} &=&
    \begin{pmatrix}
      \S_{xx}+\hat{\delta}\I_{d_x} & \0\\
      \0 & \I_{d_y}\\
    \end{pmatrix}.
\end{eqnarray*}
In a similar way to ridge regression, we can derive the GPE of CCA with $\ell_2$-norm regularization \cite{NC:Hardoon:2004,kernelICA:bach} as
\begin{align*}
  \ov{\C}^{\rm (CCA-\ell_2)} &=
    \begin{pmatrix}
      \0 & \S_{xy}\\
      \S_{yx} & \0\\
    \end{pmatrix},\\
  \ul{\C}^{\rm (CCA-\ell_2)} &=
    \begin{pmatrix}
      \S_{xx}+\hat{\delta}\I_{d_x} & \0\\
      \0 & \S_{yy}+\hat{\delta}\I_{d_y}\\
    \end{pmatrix}.
\end{align*}
In addition, we can incorporate $\ell_1$-norm regularization into the GPE framework only if the objective generalized eigenproblem has the following form:
\begin{eqnarray*}
  \X\L_{\QQ}\X^{\top}\w &=& \lambda\X\X^{\top}\w,
\end{eqnarray*}
meaning
\begin{eqnarray*}
  \S_{\QQ,xx}\w &=& \lambda\S_{xx}\w.
\end{eqnarray*}
PCA, FDA, MLR, CCA, OPLS and several variants can be included in this form. The details can be found in the previous work \cite{leastSquareFormulation:Sun}.
As shown in the above discussion, one of the major motivations that introduce the bias term of GPE is to integrate some regularization techniques within the framework of GPE.
%

%%%%
\subsection{Locality preserving projection (LPP)}
\label{sec:special:LPP}

Locality preserving projections (LPP) \cite{LPP} seeks for an embedding transformation such that nearby data pairs in the original space close in the embedding space.
Thus, LPP can reduce the dimensionality without losing the local structure.
Let $\A$ be an affinity matrix, that is, the $N$-dimensional matrix with the $(n,m)$-th element $A_{n,m}$ being the affinity between $\x_n$ and $\x_m$.
We assume that $A_{n,m}\in [0,1]$; $A_{n,m}$ is large if $\x_n$ and $\x_m$ are close and $A_{n,m}$ is small if $\x_n$ and $\x_m$ are far apart.
There are several different manners of defining $\A$, such as using the local scaling heuristics \cite{spectralClustering:zelnik}, i.e.
\begin{eqnarray*}
  A_{n,m} &=&
  \exp\left\{-\frac{\|\x_n-\x_m\|^2}{\sigma_n\sigma_m}\right\},\\
  \sigma_n &=& \|\x_n-\x_n^{(k)}\|,
\end{eqnarray*}
where $\x_n^{(k)}$ is the $k$-th nearest neighbor of $\x_n$.
A heuristic choice of $k=7$ was shown to be useful through experiments \cite{spectralClustering:zelnik}.
The objective function to be minimized is the following weighted squared error:
\begin{eqnarray*}
  \epsilon^{\rm (LPP)}(\w|\X)
  &=& \sum_{n=1}^{N}\sum_{m=1}^{N}A_{n,m}\|\w^{\top}\x_n-\w^{\top}\x_m\|^2\\
  & & \mbox{s.t. } \w^{\top}\X\D_{\A}\X^{\top}\w = 1,
\end{eqnarray*}
In the same way as the derivation of GPE (see Section \ref{sec:gpe}), the above minimization can be converted to the following generalized eigenvalue problem:
\begin{eqnarray*}
  \X\L_{\A}\X^{\top}\w &=& \lambda\X\D_{\A}\X^{\top}\w.
\end{eqnarray*}
Thus, the GPE of LPP can be obtained as
\begin{align*}
  \ov{\C}^{\rm (LPP)} &= \X\L_{\A}\X^{\top},&
  \ul{\C}^{\rm (LPP)} &= \X\D_{\A}\X^{\top}.
\end{align*}
%

%%%%
\subsection{Local Fisher discriminant analysis (LFDA)}
\label{sec:special:LFDA}

Local Fisher discriminant analysis (LFDA) \cite{LFDA_paper} is a method for supervised dimensionality reduction, and an extension of Fisher discriminant analysis (FDA).
LFDA can overcome the weakness of the original FDA against outliers.
The point is the introduction of between-sample similarity matrix $Q$ obtained from the affinity matrix, for calculating the between-class scatter matrix $\S_{\QQ}^{\rm (lb)}$ and the within-class scatter matrix $\S_{\QQ}^{\rm (lw)}$.
\begin{eqnarray*}
  \S_{\QQ}^{\rm (lb)} &=& \sum_{n=1}^{N}\sum_{m=1}^{N}Q_{n,m}^{\rm (lb)}
  (\x_n-\x_m)(\x_n-\x_m)^{\top},\\
  \S_{\QQ}^{\rm (lw)} &=& \sum_{n=1}^{N}\sum_{m=1}^{N}Q_{n,m}^{\rm (lw)}
  (\x_n-\x_m)(\x_n-\x_m)^{\top}.
\end{eqnarray*}
where $\QQ^{\rm (lb)}$ and $\QQ^{\rm (lw)}$ are the $N\times N$ matrices with
\begin{eqnarray*}
  Q_{n,m}^{\rm (lb)} &=&
    \begin{cases}
      A_{n,m}(1/N-1/N_c) & \mbox{if } y_n=y_m=c,\\
      1/N & \mbox{if } y_n\neq y_m,
    \end{cases}\\
  Q_{n,m}^{\rm (lw)} &=&
    \begin{cases}
      A_{n,m}/N_c & \mbox{if } y_n=y_m=c,\\
      1/N & \mbox{if } y_n\neq y_m,
    \end{cases}
\end{eqnarray*}
where $N_c$ is the number of samples in class $c$.
Note that the local scaling is computed in a class-wise manner in LFDA, since we want to preserve the within-class local structure.
This also contributes to reducing the computational cost for nearest neighbor search when computing the local scaling.
From the above discussion, the GPE of LFDA can be obtained as follows:
\begin{align*}
  \ov{\C}_{\QQ}^{\rm (LFDA)} &= \S_{\QQ}^{\rm (lb)}, &
  \ul{\C}_{\QQ}^{\rm (LFDA)} &= \S_{\QQ}^{\rm (lw)}.
\end{align*}
%

%%%%
\subsection{Semi-supervised LFDA (SELF)}
\label{sec:special:SELF}

Semi-supervised local fisher discriminant analysis, called SELF \cite{SELF_paper}, integrates LFDA as a supervised dimensionality reduction and PCA as a unsupervised dimensionality reduction.
SELF brings us one example for designing multivariate analysis methods via the GPE framework from the following two viewpoints:
\begin{enumerate}
\item combining several multivariate analysis methods via GPE,
\item changing sample sets to calculate the data term in GPE, which provides us to extend the method to a semi-supervised one.
\end{enumerate}
Assume that there are two samples sets $\X$ and $\Y$, each sample in $\Y$ represents a class indicator vector, and an incomplete sample set $\X^{({\rm I})}$ only exists, namely there are at least one unlabeled samples in the sample set $\X$.
In such cases, we can search for solutions that lie in the span of the larger sample set $\X$, and regularize using the additional data.
SELF looks for solutions that lie along an empirical estimate of the subspace spanned by all the samples.
This gives increased robustness to the algorithm, and increases class separability in the absence of label information.
In detail, SELF integrates the GPE ($\S_{\QQ}^{\rm (C,lb)}$ and $\S_{\QQ}^{\rm (C,lb)}$) of LFDA calculated only from the labeled samples (in other words, complete sample sets) and the GPE $\S_{xx}$ of PCA calculated from all the samples, as follows:
\begin{eqnarray*}
  \ov{\C}_{\QQ}^{\rm (SELF)} &=& \beta\S_{\QQ}^{\rm (C,lb)}+(1-\beta)\S_{xx},\\
  \ul{\C}_{\QQ}^{\rm (SELF)} &=& \beta\S_{\QQ}^{\rm (C,lw)}+(1-\beta)\I_{d_x},
\end{eqnarray*}
where $\beta$ is a hyper parameter satisfying $0\le\beta\le 1$.
When $\beta=1$, SELF is equivalent to LFDA with only the labeled samples $(\X^{({\rm C})},\Y^{({\rm C})})$.
Meanwhile, when $\beta=0$, SELF is equivalent to PCA with all samples in $\X$.
Generally speaking, SELF inherits the properties of both LFDA and PCA, and their influences can be controlled by the parameter $\beta$.
%

%%%%
\subsection{Semi-supervised CCA}
\label{sec:special:SemiCCA}

In a similar way to that of SELF, a semi-supervised extension of CCA can be derived, which is called SemiCCA \cite{SemiCCA_icpr}.
Assume that there are two samples sets $\X$ and $\Y$, and each includes incomplete sample set $\X^{({\rm I})}$ and $\Y^{({\rm I})}$ only exists, namely there are at least one unpaired samples in both $\X$ and $\Y$.
SemiCCA integrates the GPE of CCA calculated only from the complete sample sets) and the GPE of PCA calculated from the complete and incomplete sample sets, as follows:
\begin{eqnarray*}
  \lefteqn{\ov{\C}^{\rm (SemiCCA)}}\nonumber\\
  &=& \beta
  \begin{pmatrix}
    \0 & \S_{xy}^{({\rm C})}\\
    \S_{yx}^{({\rm C})} & \0\\
  \end{pmatrix}
  +(1-\beta)
  \begin{pmatrix}
    \S_{xx} & \0\\
    \0& \S_{yy}\\
  \end{pmatrix},\\
  \lefteqn{\ul{\C}^{\rm (SemiCCA)}}\nonumber\\
  &=& \beta
  \begin{pmatrix}
    \S_{xx}^{({\rm I})} & \0\\
    \0 & \S_{yy}^{({\rm I})}\\
  \end{pmatrix}
  +(1-\beta)
  \begin{pmatrix}
    \I_{d_x} & \0\\
    \0 & \I_{d_y}\\
  \end{pmatrix}
\end{eqnarray*}
When $\beta=1$, SemiCCA is equivalent to CCA with only the complete samples $(\X^{({\rm C})},\Y^{({\rm C})})$.
Meanwhile, when $\beta=0$, SemiCCA is equivalent to PCA with all samples in $\X$ and $\Y$ under the assumption that $\X$ and $\Y$ are uncorrelated with each other.
Another type of semi-supervised extension of CCA has been developed by Blaschko et al. \cite{semiSupervisedKCCA:blaschko}.
Please see the detail in Section \ref{sec:kernel}.
%

%%%%%%%%
\section{How to design new methods}
\label{sec:new}

To summarize the discussions so far, we describe (1) GPEs of major existing methods, (2) the way for integrating several GPEs and (3) some semi-supervised extensions by changing the sample sets for calculating GPEs.
This section shows that we can easily design new multivariate analysis methods at will by replicating those steps.
Note that another way to generate new methods would be possible, and the following one is only one example. 
One of the simple extensions is to integrate FDA as supervised dimensionality reduction and CCA as unsupervised dimensionality reduction with a latent model.
Consider a problem of video categorization, where its training data includes image features $\X$, audio features $\Y$ and class indexes.
Finding appropriate correlations of such three different modals would be still challenging.
Several approaches might be possible:
(1) FDA for concatenated features $(\X^{\top},\Y^{\top})^{\top}$, which cannot obtain appropriate correlations between two different types of feature vectors,
(2) CCA for two features $(\X,\Y)$ followed by FDA on the compressed domain, which cannot find class-wise differences of correlations.
Here, we newly introduce an integration of CCA and FDA, which enables us to extract class-wise differences of feature correlations as well as to achieve discriminative embedding simultaneously.
In the following, we call this method CFDA for the simplicity.
CFDA can be formulated by the following equation:
\begin{eqnarray}
  \ov{\C}_{\QQ}^{\rm (CFDA)} &=& \beta
  \begin{pmatrix}
    \0 & \S_{xy}\\
    \S_{yx} & \0\\
  \end{pmatrix}
  +(1-\beta)\S_{\QQ}^{\rm (lb)},\\
  \ul{\C}_{\QQ}^{\rm (CFDA)} &=& \beta\
  \begin{pmatrix}
    \S_{xx} & \0\\
    \0 & \S_{yy}\\
  \end{pmatrix}
  +(1-\beta)\S_{\QQ}^{\rm (lw)}.
\end{eqnarray}
When $\beta=1$ CFDA is equivalent to CCA, while when $\beta=0$ CFDA is
equivalent to FDA for concatenated features $(\X^{\top},\Y^{\top})^{\top}$.
%

%%%%%%%%
\section{Kernelized extensions}
\label{sec:kernel}

%%%%%%%%
\subsection{Kernelization of standard methods}
\label{sec:kernel:fundam}

Almost all the methods in the GPE framework can be kernelized in a similar manner to the existing ones.
First, we describe kernel CCA \cite{kernelCCA:fyfe,kernelCCA:akaho} and related regularization techniques.
The original CCA can be extended to, e.g. non-vectorial domains by defining
kernels over $\x$ and $\y$
\begin{eqnarray*}
  k_x(\x_n,\x_m) &=& \langle\phi_x(\x_n),\phi_x(\x_m)\rangle,\\
  k_y(\y_n,\y_m) &=& \langle\phi_y(\y_n),\phi_x(\y_m)\rangle,
\end{eqnarray*}
and searching for solutions that lie in the span of $\phi_x(x)$ and $\phi_y(y)$
\begin{eqnarray*}
  \w_x = \sum_{n=1}^N \alpha_n\phi_x(\x_n),\quad
  \w_y = \sum_{n=1}^N \beta_n\phi_y(\y_n).
\end{eqnarray*}
In this setting, we use the following empirical scatter matrix
\begin{eqnarray*}
  \hat{\S}_{xy} &=& \sum_{n=1}^{N}\phi_x(\x_i)\phi_y(\y_i)^{\top}.
\end{eqnarray*}
Denoting the Gram matrices defined by the samples as $\K_{xx}$ and $\K_{yy}$, the solution can be obtained from the following optimization problem with respect to coefficient vectors, $\balpha$ and $\bbeta$
\begin{eqnarray*}
  \lefteqn{\rho^{(kCCA)}(\w_x,\w_y|\X,\Y,\phi_x,\phi_y)}\nonumber\\
  \hspace{10mm}&=& \frac{\balpha^{\top}\K_{xx}\K_{yy}\bbeta}
           {\sqrt{\balpha^{\top}\K_{xx}^2\balpha \bbeta^{\top}\K_{yy}^2\bbeta}}.
\end{eqnarray*}
In the same way as CCA, the optimization can be achieved by solving the following generalized eigenvalue problem:
\begin{eqnarray*}
  \ov{\C}^{\rm (kCCA)}
  \begin{pmatrix}
    \balpha\\ \bbeta
  \end{pmatrix}
  &=& \lambda\ul{\C}^{\rm (kCCA)}
  \begin{pmatrix}
    \balpha\\ \bbeta
  \end{pmatrix}\\
  \ov{\C}^{\rm (kCCA)} &=&
    \begin{pmatrix}
      \0 & \K_{xx}\K_{yy}\\
      \K_{yy}\K_{xx} & \0
    \end{pmatrix},\\
  \ul{\C}^{\rm (kCCA)} &=&
    \begin{pmatrix}
      \K_{xx}^2 & \0\\
      \0 & \K_{yy}^2
    \end{pmatrix}.
\end{eqnarray*}
Although the bases $(\w_x,\w_y)$ cannot be obtained, the projection to those bases can be calculated with the help of the kernel trick:
\begin{eqnarray*}
  \w_x^{\top}\phi_x(\x) &=& \sum_{n=1}^N \alpha_n K_{xx}(\x,\x_n),\\
  \w_y^{\top}\phi_y(\y) &=& \sum_{n=1}^N \beta_n  K_{yy}(\y,\y_n).
\end{eqnarray*}
As discussed in \cite{NC:Hardoon:2004}, this optimization leads to degenerate solutions in the case that either $\K_{xx}$ or $\K_{yy}$ is invertible.
Therefore, the following $\ell_2$-regularized formulation should be necessary in general:
\begin{eqnarray*}
  \lefteqn{\ov{\C}^{\rm (kCCA-\ell_2)}}\nonumber\\
  &=& \begin{pmatrix}
        \0 & \K_{xx}\K_{yy}\\
        \K_{yy}\K_{xx} & \0
      \end{pmatrix},\\
  \lefteqn{\ul{\C}^{\rm (kCCA-\ell_2)}}\nonumber\\
  &=& \begin{pmatrix}
        \K_{xx}^2+\delta_x\K_{xx} & \0\\
        \0 & \K_{yy}^2+\delta_y\K_{yy}
      \end{pmatrix}.
\end{eqnarray*}
Another popular regularization technique is the graph Laplacian method \cite{graphLaplacian,semiSupervisedKCCA:blaschko}.
By using Laplacian regularization, we are able to learn directions that tend to lie along the data manifold estimated from a collection of data.
Denoting the empirical graph Laplacian $\L_x$ and $\L_y$ obtained from $\K_x$ and $\K_y$, the formulation is replaced by the following equations:
\begin{eqnarray*}
  \lefteqn{\ov{\C}^{\rm (kCCA-Lap)}}\nonumber\\
  &=& \begin{pmatrix}
        \0 & \K_{xx}\K_{yy}\\
        \K_{yy}\K_{xx} & \0
      \end{pmatrix},\\
  \lefteqn{\ul{\C}^{\rm (kCCA-Lap)}}\nonumber\\
  &=& \begin{pmatrix}
        \K_{xx}^2+\gamma_x\R_x & \0\\
        \0 & \K_{yy}^2+\gamma_y\R_y
      \end{pmatrix},\\
  \R_x &=& \K_{xx}\L_x\K_{xx},\quad \R_y = \K_{yy}\L_y\K_{yy}.
\end{eqnarray*}
%

%%%%%%%%
\subsection{Non-linear embedding methods}
\label{sec:kernel:embed}

With the kernelized extension, non-linear dimensionality reduction such as locally linear embedding \cite{LLE} and Laplacian eigenmaps \cite{laplacianEigenmap} are also in the GPE framework.
%

%%%%%%%%
\subsubsection{Laplacian eigenmaps}
\label{sec:kernel:embed:LE}

Laplacian eigenmaps \cite{laplacianEigenmap} is one of the popular methods for non-linear embedding.
The goal of Laplacian eigenmaps is to find an embedding that preserves the local structure of nearby high-dimensional samples.
Laplacian eigenmaps exploits graph Laplacian of a neighboring graph on the samples $\X$, where each edge measures the affinity between two samples.
Since a set of edge weights can be expressed by a Gram matrix $K_x$, the objective function of Laplacian eigenmaps to be minimized is
\begin{eqnarray*}
  \rho^{\rm (LE)}(\w_x|\X)
  &=& (\balpha^{\top}\L_x\balpha)(\balpha^{\top}\D_x\balpha)^{-1},
\end{eqnarray*}
where $\D_x$ is a diagonal matrix satisfying $\D_x=\K_x+\L_x$.
Therefore, the GPE of Laplacian eigenmaps can be obtained as
\begin{align*}
  \ov{\C}^{\rm (LE)} &= \L_x,& \ul{\C}^{\rm (LE)} &= \D_x.
\end{align*}
%

%%%%%%%%
\subsubsection{Local linear embedding (LLE)}
\label{sec:kernel:embed:LLE}

Local linear embedding (LLE) \cite{LLE} finds an embedding of the samples $\X$ that preserves the local structure of nearby samples in the high-dimensional space.
LLE builds the embedding by preserving the geometry of pairwise relations between samples in the high-dimensional manifold.
LLE first computes a Gram matrix $\K_x$ containing the structural information of the embedding by minimizing the following function:
\begin{eqnarray*}
  \rho^{\rm (LLE1)}(\K_x|\X) &=& \|\X(\I_N-\K_x)\|_F^2\\
  && \mbox{s.t. } \K_x\1_N = \1_N,
\end{eqnarray*}
where each column of the Gram matrix $\K_x$ has $k$ non-zero values.
This minimization can be solved via a linear system of equations.
Once $\K_x$ is calculated, LLE next finds a base that minimizes
\begin{eqnarray*}
  \rho^{\rm (LLE2)}(\w_x|\X)
  &=& \balpha^{\top}(\I_N-\K_x)(\I_N-\K_x)^{\top}\balpha,\\
  & & \mbox{s.t. } \balpha^{\top}\balpha=1.
\end{eqnarray*}
Therefore, the GPE of LLE can be obtained as
\begin{eqnarray*}
  \ov{\C}^{\rm (LLE)} &=& (\I_N-\K_x)(\I_N-\K_x)^{\top},\\
  \ul{\C}^{\rm (LLE)} &=& \I_N.
\end{eqnarray*}
%

%%%%%%%%
\subsection{Clustering methods}
\label{sec:kernel:cluster}

With the benefit of the kernelized extension of the GPE framework, several clustering methods such as spectral clustering (SC) \cite{spectralClustering,multiclassSpectralClustering,correlationalSpectralClustering} and normalized cuts (NC) \cite{normalizedCuts} are also in the class of multivariate analysis we are concerned with.
\begin{align*}
  \ov{\C}^{\rm (SC)} &= \L_x, & \ul{\C}^{\rm (SC)} &= \D_x,\\
  \ov{\C}^{\rm (NC)} &= \D_x^{-1/2}\L_x\D_x^{-1/2}, & \ul{\C}^{\rm (NC)} &= \I_N.
\end{align*}
Kernel k-means \cite{kernelKMeans} is also known to belong to this family if we admit to introduce a certain iterative procedure \cite{hardProbClustering}. The details can be seen in the preceding work by De la Torre \cite{DelaTorre2010,ls-wkrrrPAMI}.
%

%%%%%%%%
\subsection{How to design new kernelized methods}
\label{sec:kernel:new}

Integrating two methods within the kernelized GPE framework is not obvious, since a simple addition of Gram matrices is not GPE.
One example can be seen in a kernelized extension of SELF, called kernel SELF \cite{SELF_paper}.
Remember that the original SELF integrates LFDA with labeled samples and PCA with all the samples (see Section \ref{sec:special:SELF}), and it can be formulated by a localized between-class scatter matrix $\S_{\QQ}^{\rm (C,lb)}$, localized within-class matrix $\S_{\QQ}^{\rm (C,lw)}$ and the ordinary scatter matrix $\S_{xx}$.
Kernel SELF can be formulated via their Laplacian matrices $\L_{\QQ}^{\rm (C,lb)}$, $\L_{\QQ}^{\rm (C,lw)}$, $\L_{xx}$, as follows:
\begin{eqnarray*}
  \ov{\C}^{\rm (kSELF)}
  &=& \K_x\{\beta\L_{\QQ}^{\rm (C,lb)}+(1-\beta)\L_{xx}\}\K_x,\\
  \ul{\C}^{\rm (kSELF)}
  &=& \beta\K_x\L_{\QQ}^{\rm (C,lb)}\K_x+(1-\beta)\K_x.
\end{eqnarray*}
From this formulation, we can see that a weighted sum of GPEs in original multivariate analysis corresponds to a weighted sum of Laplacian matrices in
kernelized multivariate analysis.
Namely, when dealing with kernelized multivariate analysis, we have to explicitly derive GPEs of existing methods, and replace the data matrix into its Gram matrix.
%

%%%%%%%%
\section{Concluding remarks}
\label{sec:conclude}

This paper provided a new theoretical expression of covariance matrices and Gram matrices, which we call generalized pairwise expression (GPE).
This provided a unified insight into various multivariate analysis methods and their extensions.
GPE made it easy to design desired multivariate analysis methods by simple
combinations of GPEs of existing methods as templates.
According to this methodology, we designed several new multivariate analysis methods.
The GPE framework covers a wide variety of multivariate analysis methods, and thus the way we have presented in this paper for designing new methods is still one of the examples. Developing more general guidelines would be significant future work.
%

%%%%%%%%
\bibliographystyle{ieicetr}
\bibliography{gpe}

\end{document}